\documentclass{article}
\usepackage{spconf,amsmath,graphicx,hyperref}
\usepackage{verbatim}
\usepackage{booktabs}       % professional-quality tables
\usepackage{amsfonts}       % blackboard math symbols
\usepackage{nicefrac}       % compact symbols for 1/2, etc.
\usepackage{microtype}      % microtypography
\usepackage{xcolor}         % colors
\usepackage{multirow}
\usepackage{color}
\usepackage{caption}
\usepackage{wrapfig}

\title{Intra-Class Subdivision for Pixel Contrastive Learning: Application to
Semi-supervised Cardiac Image Segmentation}
\name{Jiajun Zhao, Xuan Yang 
\thanks{Thanks to the Shenzhen Fundamental Research Program (JCYJ20220531102407018), Guangdong Province Key Laboratory of Popular High-Performance Computers 2017B030314073, and National Natural Science Foundation of China (Grant No.62201355). Corresponding author email:  yangxuan@szu.edu.cn}}
\address{College of Computer Science and Software Engineering, Shenzhen University, Shenzhen 518060, China}

\begin{document}
\topmargin=0mm
\ninept
\maketitle
\begin{abstract}
We propose an intra-class subdivision pixel contrastive learning (SPCL) framework for cardiac image segmentation to address representation contamination at boundaries. The novel concept ``Unconcerned sample'' is proposed to distinguish pixel representations at the inner and boundary regions within the same class, facilitating a clearer characterization of intra-class variations. A novel boundary contrastive loss for boundary representations is proposed to enhance representation discrimination across boundaries. The advantages of the unconcerned sample and boundary contrastive loss are analyzed theoretically. Experimental results in public cardiac datasets demonstrate that SPCL significantly improves segmentation performance, outperforming existing methods with respect to segmentation quality and boundary precision. Our code is available at \url{https://github.com/Jrstud203/SPCL}.
\end{abstract}
\section{Introduction}
\label{sec:intro}
Contrastive learning (CL) has attracted considerable attention in semi-supervised learning to exploit the semantic relationships between samples and endow the segmentation model with a powerful feature extraction capability~\cite{sung2024contextrast,wang2022uncertainty,wang2023hunting,hu2021region,basak2023pseudo}. It is applicable to medical image segmentation due to the scarcity of annotated data~\cite{lin2025enhancing,tang2024semi,zhou2024multi}.
The underlying idea of CL is to pull together samples from the same class while pushing apart negative samples in feature space, thus imposing intra-class compactness and inter-class separation. 
Contrastive learning is challenged by intra-class feature diversity and inter-class feature ambiguity, especially near object boundaries; 
approaches such as hard negative mining~\cite{sung2024contextrast,xia2022progcl} and boundary-specific strategies~\cite{yang2025boundary,zhang2024boundaryaware} have been proposed to mitigate this issue.

\begin{figure}[t]
\centering
\includegraphics[width=0.95\linewidth]{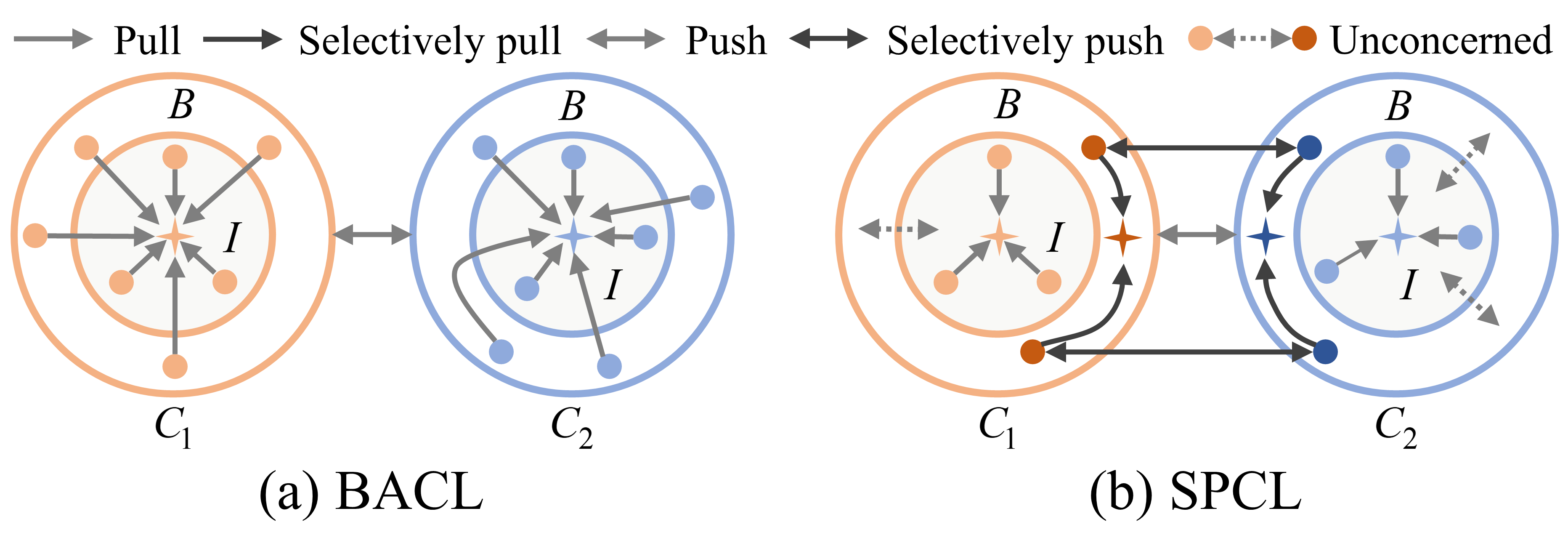}
\caption{Comparison of existing BACL versus our SPCL. BACL methods pull both inner and boundary samples toward the same centroid, while our SPCL aligns them to separate centroids, addressing intra-class variations and enabling both to act as negatives in inter-class contrastive learning.}
\label{moti}
\end{figure}

Some boundary-aware CL (BACL) methods have been proposed to mitigate this issue. For example, CBL~\cite{wu2023conditional} and BUS~\cite{choe2024open} ignore long-range pixels and perform CL in a local structure, by pulling an anchor closer to its neighbors of the same class and pushing it away from its neighbors of different classes. BoCLIS~\cite{yang2025boundary} introduces a patch sampling strategy to alleviate intra-class pixel feature differences, maximizing the similarities between the class prototype patch and the intra-class boundary and inner patches. To avoid the impact of boundary features on inner features, BASS~\cite{zhang2024boundaryaware} performs region-based contrastive learning separately within the boundary and inner regions.
However, boundary features incorporate information from multiple classes, exhibiting substantial distinctions with inner features; the above methods still assign them to the same class and attempt to pull them closer, causing conflicts or difficulties in representation optimization, as shown in Fig.~\ref{moti} (a). Furthermore, although BASS treats the inner and boundary regions individually to deal with intra-class variance, intra-class samples from both regions do not simultaneously appear as negative samples in inter-class contrastive learning. Importantly, these methods overlook the discrepancy of boundary features, which degrades segmentation accuracy.

Previous research has shown that partitioning semantic categories into finer subclasses can enhance prediction performance~\cite{guan2022unbiased,li2023semantic}. However, existing subclass-partitioned segmentation approaches have not been extended to contrastive frameworks that explicitly model inter-subclass relationships. To address this gap, we propose to integrate subclass partitioning into the feature space of contrastive learning, where distinct contrastive schemes are applied to different subclasses in order to fully exploit the fine-grained structure of the feature space. Our approach reduces intra-class variance and strengthens inter-class discriminability, as it avoids indiscriminately clustering heterogeneous samples within the same class.

Building on this insight, we propose an intra-class Subdivision Pixel Contrastive Learning (SPCL) framework for semi-supervised cardiac image segmentation. The core idea of SPCL is to leverage the underlying geometry of the feature space to model intra-class heterogeneity, distinguishing inner pixels from boundary pixels as separate subclasses to promote more discriminative feature learning. 
To maintain the subtle relationship between subclasses, we introduce a novel concept of the ``unconcerned sample'' in SPCL. As shown in Fig.~\ref{moti}(b), for an inner/boundary anchor, unconcerned samples refer to the boundary/inner pixels of the same class. Although unconcerned samples belong to the same class, they are not considered positive samples for the anchor due to intra-class variance. Instead, both serve as negative samples for the anchors from other classes to ensure inter-class separability. 
Furthermore, we theoretically demonstrate that the proposed unconcerned sample enhances the separability of pixel representations.
To improve the compactness of intra-subclass and the separability of inter-class for inner pixels, we introduce inner contrastive learning (ICL) with unconcerned samples to pull the inner anchor closer to positive samples from the same subclass and push it away from negative samples from different classes. Additionally, we design a novel boundary-based contrastive loss (BCL) to enhance the inter-subclass separability of boundary pixels and achieve more accurate segmentation at the boundaries. We validate the effectiveness of SPCL on public cardiac datasets in various semi-supervised settings, achieving state-of-the-art segmentation performance. 

\section{Method}

\begin{figure}[t]
\centering
\includegraphics[width=0.95\linewidth]{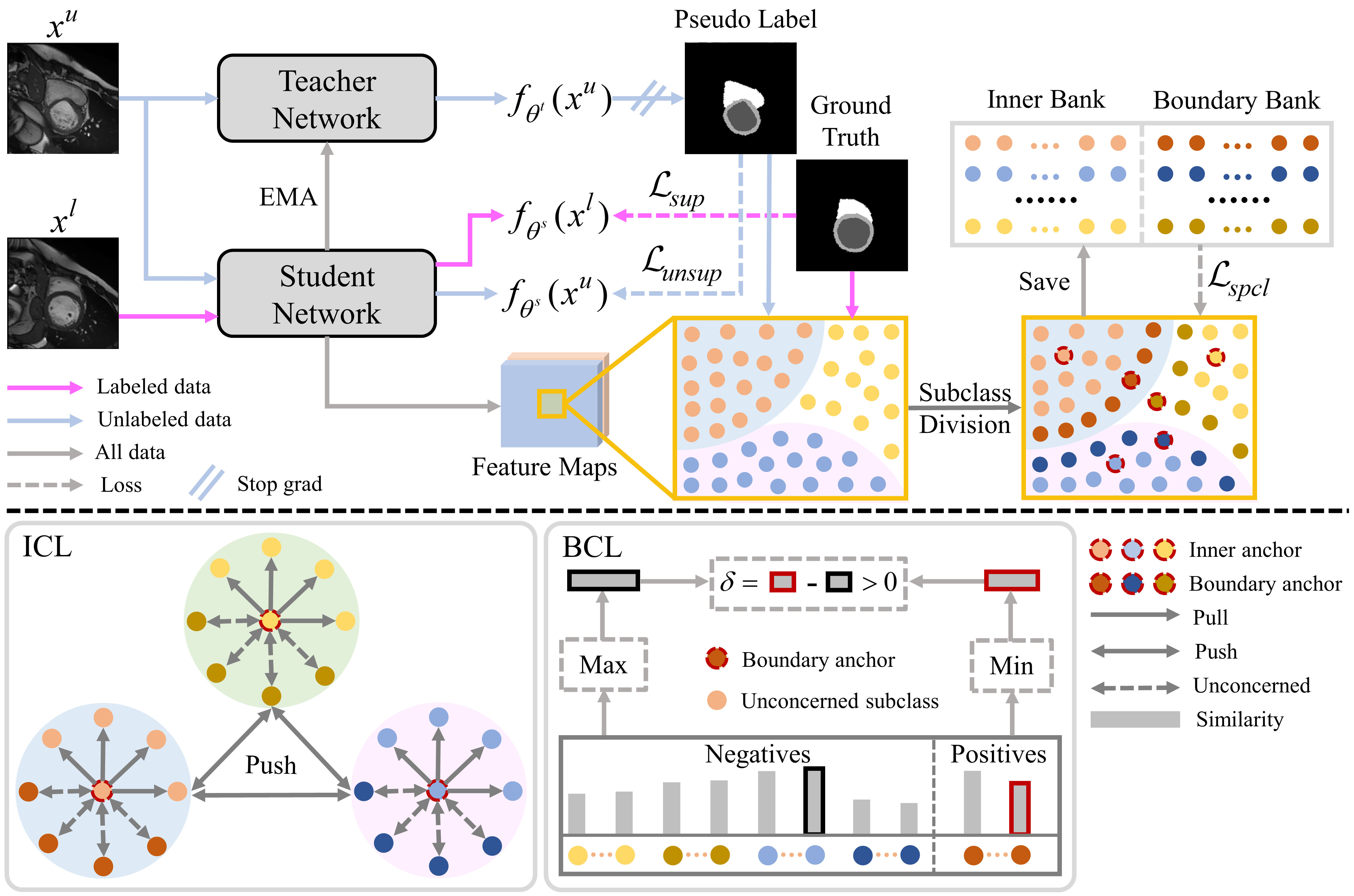}
\caption{Overview of our SPCL method built on a teacher-student network. Based on the pseudo labels of unlabeled images and the ground truth of labeled images, the representations within the feature maps are categorized into inner and boundary subclasses. Subsequently, inner and boundary contrastive learning are applied to learn the representations of the inner and boundary regions, respectively.}
\label{frame}
\end{figure}
Fig.~\ref{frame} illustrates an overview of our SPCL framework, which is built on a teacher-student network~\cite{tarvainen2017mean}. The student network $f_{\theta^s}$ is trained by gradient descent, while the teacher network $f_{\theta^t}$ is updated through the exponential moving average (EMA) of the student network. 
Given a collection of small label data $D_L=\{(x^l_i,y^l_i)^N_{i=1}\}$ and massive unlabeled images $D_U=\{(x^u_i)^M_{i=1}\}$, $N\ll M$,
our SPCL involves two training strategies: a) Label-guided training to learn pixel-level category assignment; 
b) Subclass-guided pixel contrastive learning to regularize the structures of clusters in feature space.

For label-guided training, labeled and unlabeled images are fed into the student network to predict probabilities $p_s^l = f_{\theta^s}(x^l)$ and $p_s^u = f_{\theta^s}(x^u)$, respectively. For unlabeled images, the predictions of the teacher network are used as pseudo labels $y_{t}^u=\arg\max_cp_{t}^u$, where $p_t^u = f_{\theta^t}(x^u)$. With the supervised and unsupervised losses, the student network is supervised by ground truth and pseudo labels (see Eq.~\eqref{loss}). 
For subclass-guided pixel contrastive learning, the student network outputs projection feature maps $Z$ of input images, and the pixels in the feature maps are categorized into inner and boundary subclasses, as detailed in Section~\ref{sd}. A reliable-aware threshold sampling strategy is introduced to select anchor, positive, and negative samples (see Section~\ref{sam}). The inner and boundary contrastive learning losses are explained in Section~\ref{cl}.

\subsection{Subclass Division and Unconcerned Sample} \label{sd}
Our SPCL method separates the inner and boundary pixels of the same class into distinct subclasses. 
Formally, the boundary regions are obtained through the Canny operator: $\{B^{c}\}_{c=1}^{C} = \text{Canny}(y)$, where \( C \) is the number of classes, and $y$ is the ground truth $y^l$ and the pseudo label $y_t^u$ for labeled and unlabeled images, respectively. $B^{c}$ denotes the binary boundary mask of class $c$.
The corresponding inner mask $I^{c}=y^c - B^c$, where $y^c$ is the mask of class $c$. The inner pixels of each class $c$ are assigned to the subclass \( \hat{C}_{c,I} \), while the boundary pixels are assigned to the subclass \( \hat{C}_{c,B} \). The total number of subclasses \( \hat{C} \) is \( 2C \).

We propose a novel concept ``unconcerned sample'' to preserve the subtle relationship between subclasses simply yet effectively. Specifically, unconcerned samples for inner anchors of subclass $\hat{C}_{c,I}$ are boundary pixels from subclass $\hat{C}_{c,B}$; conversely, for boundary anchors of subclass $\hat{C}_{c,B}$, they are inner pixels from subclass $\hat{C}_{c,I}$.
In contrastive learning, unconcerned samples are neither positive nor negative for anchors from the same class \( \hat{C}_{c} \), but negative for anchors from all other classes.
The main purpose of unconcerned samples is to reduce intra-class feature variations, facilitating the aggregation of inner features and encouraging inter-class separation.

\subsection{Reliable-aware Threshold Sampling} \label{sam}

Effective sampling of positive and negative pairs is of utmost importance for our SPCL. For unlabeled images, we are confident in pixel features with reliable pseudo-labels. 
Notably, anchors are chosen from the teacher-provided pseudo-labels by selecting samples that the student model finds difficult to predict (hard samples), as these hard samples offer greater learning value.
For each inner/boundary subclass, we sample a fixed number of hard samples as anchors:
\begin{equation}
\label{anchor}
\mathcal{A}^{c,id}=\{Z^{c,id}\mid \max(p_{s}) \leq  \gamma_s \land \max(p) \geq  \gamma_t\},\; id\in\{I,B\}
\end{equation}
where $p_{s}=\{p^l_{s}, p^u_{s}\}$, $p=\{y^l, p_{t}^u \}$, and $\gamma_s=0.75$ along with $\gamma_t=0.95$ are confidence thresholds. The term $\max(p) \geq  \gamma_t$ ensures that the pseudo-label is confident, and $\max(p_{s}) \leq  \gamma_s$ aims to select hard samples. $Z^{c,id}$ represents the feature subset of the \( c \)-th inner/boundary subclass for the projection feature map generated by the student model.

Due to the long-tail phenomenon in the cardiac dataset, negative samples for specific classes (e.g., myocardium and right ventricular) may be extremely limited within a mini-batch.  
To address this, we use a subclass-wise inner/boundary memory bank \( \mathcal{Q}^{I/B}_c \) to store the centroid \(\bar{z}^{c,id} = \frac{1}{\left | S_i^{c,id} \right |} \sum_{z_i^{c,id} \in S_i^{c,id}} z_i^{c,id} \cdot \mathbf{1}(\max(p) \geq  \gamma_t )\) for each inner/boundary subclass \( \hat{C}_{c,id} \) on a given map, where \( S_i^{c,id} \) is the feature set of the \( c \)-th inner/boundary subclass for the $i$-th projection feature map. The memory bank \( \mathcal{Q}^{I/B}_c \) is updated FIFO while maintaining a fixed size.

\subsection{Inner and Boundary Contrastive Learning}  \label{cl}
\textbf{Inner Contrastive Learning.}
ICL aims to explicitly strengthen the intra-subclass compactness and inter-class separation for inner features by increasing the similarities of positive pairs and reducing the similarities of negative pairs. We adopt the commonly-used InfoNCE~\cite{oord2018representation} loss considering unconcerned samples, which is formulated as:
\setlength{\abovedisplayskip}{2pt}
\setlength{\belowdisplayskip}{2pt}
\begin{equation}
\begin{split}
\label{icl}
\mathcal{L}_{icl} = &-\frac{1}{\left | \mathcal{A}^I \right | } {\sum_{i\in \mathcal{A}^I}} \frac{1}{\left | P_i \right | } \sum_{z^+_i \in P_i}
\log \frac{e^{z_i \cdot z^+_i/\tau }}{e^{z_i \cdot z^+_i/\tau } \!+\! { \textstyle \sum _{z_i^- \in N_i} e^{z_i \cdot z_i^-/\tau }} } ,\\
\end{split}
\end{equation}
where $\mathcal{A}^I=\cup_{c=1}^C \{ \mathcal{A}^{c,I} \}$ is the set of inner anchors from the current mini-batch, $P_i$ and $N_i$ represent positive and negative sample sets for the $i$-th inner anchor. Note that these sample sets are extracted from the inner and boundary memory banks \( \mathcal{Q}^{I/B} \). $z_i$ is the feature of the $i$-th inner anchor from a subclass $\hat{C}_{c,I}$, $z^+_i$ is a positive feature of the same subclass, and $z^-_i$ is a negative feature from different subclasses other than the unconcerned subclass $\hat{C}_{c,B}$. 
Importantly, in $\mathcal{L}_{icl}$, unconcerned samples serve neither as positive nor as negative samples to the inner anchor $z_i$.
$\tau=0.5$ is a temperature hyper-parameter.

To analyze the influence of unconcerned samples on $\mathcal{L}_{icl}$ for an inner anchor $z_i$, we
denote the positive similarity and negative similarity as $\Upsilon^+_i=z_i \cdot z^+_i$ and $\Upsilon^-_i=z_i \cdot z^-_i$, respectively. The ICL loss is approximated as:
\begin{equation}
{\scriptsize
\begin{split}
\label{icl_zi}
\mathcal{L}_{icl}(z_i) 
\approx &\log \left[1+ e^{- \Upsilon^+_i/\tau} (\lvert N_i \rvert -1) \mathbb {E}_{\Upsilon^-_i}( e^{\Upsilon^-_i /\tau } ) \right]\\
\ge &\log \left[ e^{- \Upsilon^+_i/\tau} (\lvert N_i \rvert -1) \mathbb {E}_{\Upsilon^-_i}( e^{\Upsilon^-_i /\tau } ) \right]\\
\ge & {- \Upsilon^+_i/\tau}+  \log(\lvert N_i \rvert -1) + \mathbb {E}_{\Upsilon^-_i} {\Upsilon^-_i /\tau }.  \\
\end{split}
}
\end{equation}

\begin{wrapfigure}{l}{0.5\linewidth}  
    \includegraphics[width=1\linewidth]{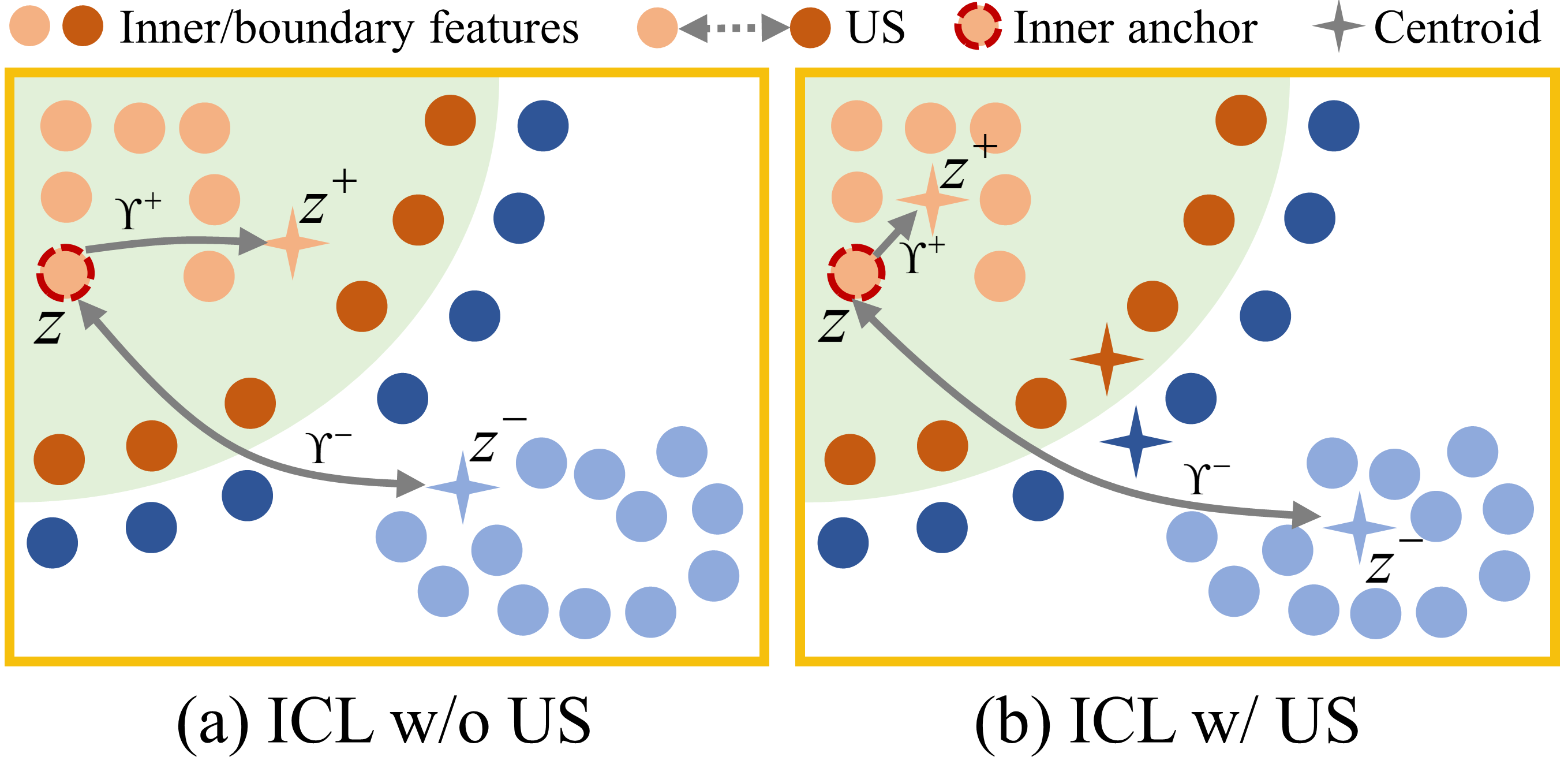}
    \caption{Influence of unconcerned samples (US) on ICL.}
\label{ficl}
\vspace{-0.2cm}
\end{wrapfigure}

Fig.~\ref{ficl} illustrates the impact of the unconcerned sample on the ICL loss, where inner features generally exhibit greater compactness compared with boundary features. Without considering unconcerned samples, the mixed centroid of inner and boundary features lies between them, resulting in a greater distance and smaller positive similarity, while negative similarity tends to be larger. By contrast, when unconcerned samples are incorporated, inner and boundary centroids are confined to their respective regions, leading to greater positive similarity and smaller negative similarity.
Therefore, we can conclude that the lower bound of the InfoNCE loss for ICL becomes tighter when unconcerned samples are considered. 
It implies that our $\mathcal{L}_{icl}$ can make the intra-class features more compact and the inter-class features easier to distinguish.

\textbf{Boundary Contrastive Learning.}
Since boundary features are formed by a mixture of representations from different objects, they exhibit significant feature discrepancies. BCL is proposed to encourage the similarities between the boundary anchor and positive samples to be higher than those between the anchor and negative samples. 
Given a boundary anchor from the subclass \( \hat{C}_{c, B} \), we first exclude unconcerned samples from the subclass \( \hat{C}_{c, I} \). Positive samples are selected from \( \mathcal{Q}^{B}_c \), while negative samples are chosen from the remaining samples in \( \mathcal{Q}_{-c}^{I/B}= \cup_{c=1}^C  \mathcal{Q}_c^{I/B} \setminus \mathcal{Q}_{c}^{I/B}\).

The objective of BCL is to ensure that
$
\min_{z^+_i \in P_i}(\Upsilon^+_i)>\max_{z^-_i \in N_i}(\Upsilon^-_i) 
$ for the $i$-th boundary anchor.
To this end, a boundary-based contrastive loss function is formulated through existing functional approximations~\cite{dugas2000incorporating,nielsen2016guaranteed,qiao2023fuzzy}:
\setlength{\abovedisplayskip}{2pt}
\setlength{\belowdisplayskip}{2pt}
{\scriptsize
\begin{equation}\label{bcl0}
\begin{split}
\mathcal{L}_{bcl}&= \frac{1}{\left | \mathcal{A}^B \right | } {\sum_{i\in \mathcal{A}^B}} \text{ReLU}(\max_{z^-_i \in N_i}(\Upsilon^-_i) - \min_{z^+_i \in P_i}(\Upsilon^+_i) ) \\
&\approx \frac{1}{\left | \mathcal{A}^B \right | } {\sum_{i\in \mathcal{A}^B}} \text{ReLU} (\log{{\textstyle \sum\nolimits_{z^-_i \in N_i}e^{\Upsilon^-_i}  }} + \log{{\textstyle \sum\nolimits_{z^+_i \in P_i}e^{-\Upsilon^+_i}  }}) \\
&\approx \frac{1}{\left | \mathcal{A}^B \right | } {\sum_{i\in \mathcal{A}^B}} \log(1+{\textstyle \sum\nolimits_{z^+_i \in P_i}e^{-\Upsilon^+_i}  } \times {\textstyle \sum\nolimits_{z^-_i \in N_i}e^{\Upsilon^-_i}  }),
\end{split}
\end{equation}
}
where $\mathcal{A}^B=\cup_{c=1}^C \{ \mathcal{A}^{c,B} \}$ is the set of boundary anchors.
Denote $sr_i^+={\scriptstyle\sum_{z^+_i \in P_i}e^{-\Upsilon^+_i}  }$ and $sr_i^-= {\scriptstyle\sum_{z^-_i \in N_i}e^{\Upsilon^-_i}  }$ for simplity.
The gradients of $\mathcal{L}_{bcl}$ with respect to the similarities of positive and negative pairs are
$
\frac{\partial \mathcal{L}_{bcl}}{\partial \Upsilon^+_i}\!=\!-\frac{ {sr}_i^-  e^{-\Upsilon^+_i}}{1\!+{sr}_i^+  {sr}_i^-} $ and $
\frac{\partial \mathcal{L}_{bcl}}{\partial \Upsilon^-_i}\!=\!\frac{{sr}_i^+  e^{\Upsilon^-_i}}{1\!+\! {sr}_i^+ {sr}_i^-}
$, respectively.
We can see that $\left\lvert{\frac{\partial \mathcal{L}_{bcl}}{\partial \Upsilon^+_i}}/{\frac{\partial \mathcal{L}_{bcl}}{\partial \Upsilon^-_i}} \right \rvert = \frac{{e^{-\Upsilon^+_i}}/ {{sr}_i^+ }}{{e^{\Upsilon^-_i}}/ {{sr}_i^- }}$ decreases with increasing similarity. 
It implies that the relative penalty concentrates more on the less similar region, which is beneficial to push representations of boundary pixels to be similar to each other. 

Compared to InfoNCE, $\mathcal{L}_{bcl}$ is less “greedy” due to the lack of pairwise samples considered. It retains the right relative connection between positive and negative samples. In contrast, InfoNCE focuses on gathering as many positive samples as possible, which is not applicable to the case of limited boundary samples. 
The advantage of $\mathcal{L}_{bcl}$ is that it can distort the feature space to accommodate outliers and adjust to intra-class variance for distinct classes, especially applicable to the boundary pixels with mixed representations of distinct classes.

Finally, the student network is optimized by a unified loss:
\begin{equation}
\label{loss}
\begin{split}
\mathcal{L}&=\mathcal{L}_{sup} + \lambda_u \mathcal{L}_{unsup} +  \lambda_c \mathcal{L}_{spcl}\\
\mathcal{L}_{spcl}&= (1 - \alpha) \mathcal{L}_{icl} + \alpha \mathcal{L}_{bcl},\\
\end{split}
\end{equation}
where $\lambda_u=0.3$ and $\lambda_c=0.1$ are regularization parameters to balance losses. Cross-entropy and dice losses are used as $\mathcal{L}_{sup}=(\mathcal{L}_{ce}(p_s^l,y^l)+\mathcal{L}_{dice}(p_s^l,y^l))/2$. Dice loss is used as $\mathcal{L}_{unsup}=\mathbf{1}_{\left\{\max(p_{t}^u) \geq \gamma_t \right\}}\mathcal{L}_{dice}(p_{s}^u,y_{t}^u)$.
The temperature parameter $\alpha$ is increased from 0 to 0.1 during training.
The teacher network is updated as the EMA of the student network, with a momentum parameter $\beta=0.999$, $\theta^t= \beta \theta^t + (1-\beta) \theta^s$.

% total results
\begin{table*}[htp]
  \caption{Comparison of different methods on SCD, ACDC and M\&Ms datasets across different semi-supervised settings. \textbf{Bold} denotes the best performance, while \underline{underlined} denotes the second-best performance.}
  \scriptsize
  \label{whole-result}
  \centering
  \renewcommand{\arraystretch}{1.2} 
  \setlength{\tabcolsep}{1mm}{ 
\begin{tabular}{l|ccc|ccc|ccc|ccc|ccc|ccc}
\bottomrule
\multirow{2}{*}{Method} & \multicolumn{3}{c|}{SCD$_{10\%}$} & \multicolumn{3}{c|}{SCD$_{20\%}$} & \multicolumn{3}{c|}{ACDC$_{5\%}$} & \multicolumn{3}{c|}{ACDC$_{10\%}$} & \multicolumn{3}{c|}{M\&Ms$_{2.5\%}$} & \multicolumn{3}{c}{M\&Ms$_{5\%}$}\\
\cline{2-19}      & Dice$\uparrow$  & Jacc$\uparrow$ & HD95$\downarrow$  & Dice$\uparrow$  & Jacc$\uparrow$ & HD95$\downarrow$  & Dice$\uparrow$  & Jacc$\uparrow$ & HD95$\downarrow$  & Dice$\uparrow$  & Jacc$\uparrow$ & HD95$\downarrow$ & Dice$\uparrow$  & Jacc$\uparrow$ & HD95$\downarrow$  & Dice$\uparrow$  & Jacc$\uparrow$ & HD95$\downarrow$ \\
\hline
SupOnly &   84.27 & 74.35 & 8.43  & 87.08 & 78.47 & 8.44  & 77.52 & 65.38 & 26.87 & 85.05 & 74.92 & 7.59 & 70.99 & 57.61 & 19.05 & 77.57 & 65.30 & 8.73 \\
\hline
UGPCL~\cite{wang2022uncertainty} & 86.78 & 77.52 & 7.65  & 88.36 & 79.99 & 7.93  & 84.76 & 75.01 & 9.68  & 88.62 & 80.53 & 2.97 & 81.41 & 70.06 & 4.67  & 82.46 & 71.97 & 3.76 \\
DGCL~\cite{wang2023hunting}  & 87.47 & 78.71 & 4.27  &  89.36 & 81.45 & 2.34  & 86.54 & 76.92 & 3.49  & 89.42 & 81.26 & 3.12 & 82.12 & 70.80 & 3.35  & 85.16 & 75.02 & 3.11\\
ADMT~\cite{zhao2024alternate}  & 87.29 & 78.68 & 4.83  & 88.29 & 79.85 & 4.88  & 87.00 & 77.88 & 3.28  & 89.27 & 81.07 & \underline{1.92} & 82.49 & 71.36 & 4.86  & 84.62 & 74.30 & 4.37 \\
ABD~\cite{chi2024adaptive}   & 88.23 & 79.41 & 6.25  & 90.70 & 83.58 & 3.63  & 87.19 & 77.97 & \underline{3.25}  & \underline{89.47} & 81.36 & 3.69 & 84.12 & 73.47 & 4.96  & 85.00 & 74.81 & 3.82 \\
BUS~\cite{choe2024open}   & 88.07  & 79.95 & 5.69  & 90.75  & 83.55 & 2.76  & 87.17 & 77.69 & 5.87  & 89.11 & 80.85 & 2.76 & 83.50 & 72.67  & 2.74  & 85.87 & 76.02 & 2.19\\
BASS~\cite{zhang2024boundaryaware}  & \underline{90.37} & \underline{83.14}  & \underline{2.43}  &  \underline{91.57} & \underline{84.91} & \underline{1.94}  & \underline{87.60} & \underline{78.84} & 3.74  & 89.38 & \underline{81.84} & 2.24 & \underline{84.39} & \underline{73.97} & \underline{2.28}  & \underline{86.10} & \underline{76.46} & \underline{2.15} \\
\hline
SPCL (\textbf{Ours}) &  \textbf{91.34} & \textbf{84.42} & \textbf{1.62} & \textbf{92.58} & \textbf{86.41} & \textbf{1.43} & \textbf{88.66} & \textbf{80.21} & \textbf{2.81} & \textbf{90.27} & \textbf{82.76} & \textbf{1.64} & \textbf{85.36} & \textbf{75.26} & \textbf{2.12} & \textbf{87.04} & \textbf{77.70} & \textbf{2.11}\\
\toprule
\end{tabular}%
\vspace{-10pt}
}
\end{table*}

\section{Experiments} \label{exp}
\subsection{Datasets and Implementation Details} 
We evaluated our proposed SPCL on three public cardiac datasets: SCD~\cite{radau2009evaluation}, ACDC~\cite{bernard2018deep} and M\&Ms~\cite{campello2021multi}.
The SCD dataset comprises 90 short-axis cine MRI scans from 45 patients, with 30 patients used for training and 15 for testing. For semi-supervised experiments, scans from 3 (10\%) and 6 (20\%) patients in the training set were labeled. The ACDC dataset contains 200 short-axis cine MRI scans from 100 patients. Following~\cite{wang2022uncertainty}, we adopted 70 patients for training and 30 for testing, with 3 (5\%) and 7 (10\%) patients labeled in the training set. The M\&Ms dataset includes 640 short-axis annotated scans from 320 subjects, split into 170 subjects for training and 150 for testing. Scans from 4 (2.5\%) and 8 (5\%) subjects were labeled in the training set.
The Dice coefficient (Dice), the Jaccard index (Jacc), and the Hausdorff distance 95 (HD95) are used to evaluate segmentation performance.

\subsection{Quantitative Comparison}
\begin{figure}[htb]
\begin{minipage}[b]{0.6\linewidth}
  \centering
  \includegraphics[width=5.2cm]{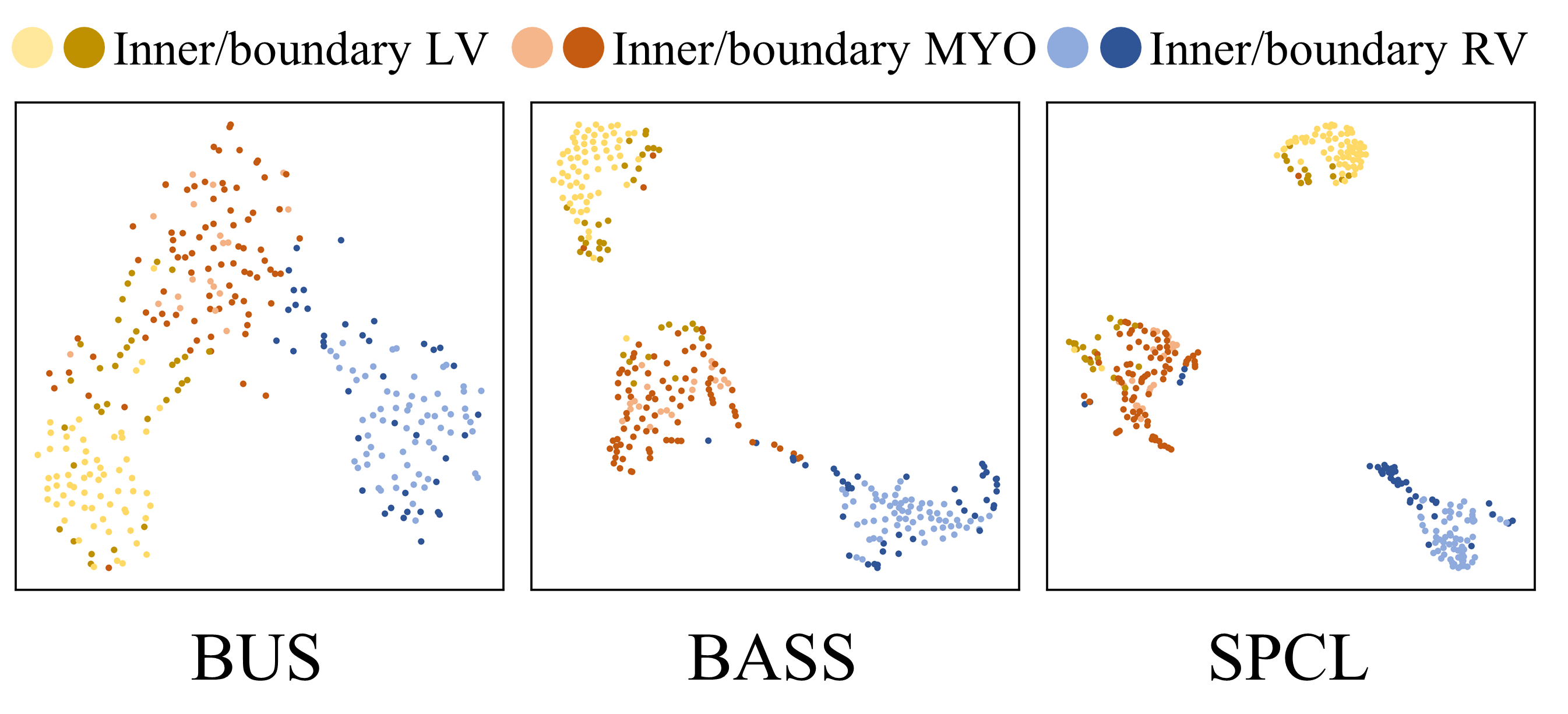}
  \caption{Visualization of feature space obtained by different BACL methods.}\label{tsne}
  \medskip
\end{minipage}
\hfill
\begin{minipage}[b]{0.36\linewidth}
  \centering
  \includegraphics[width=3cm]{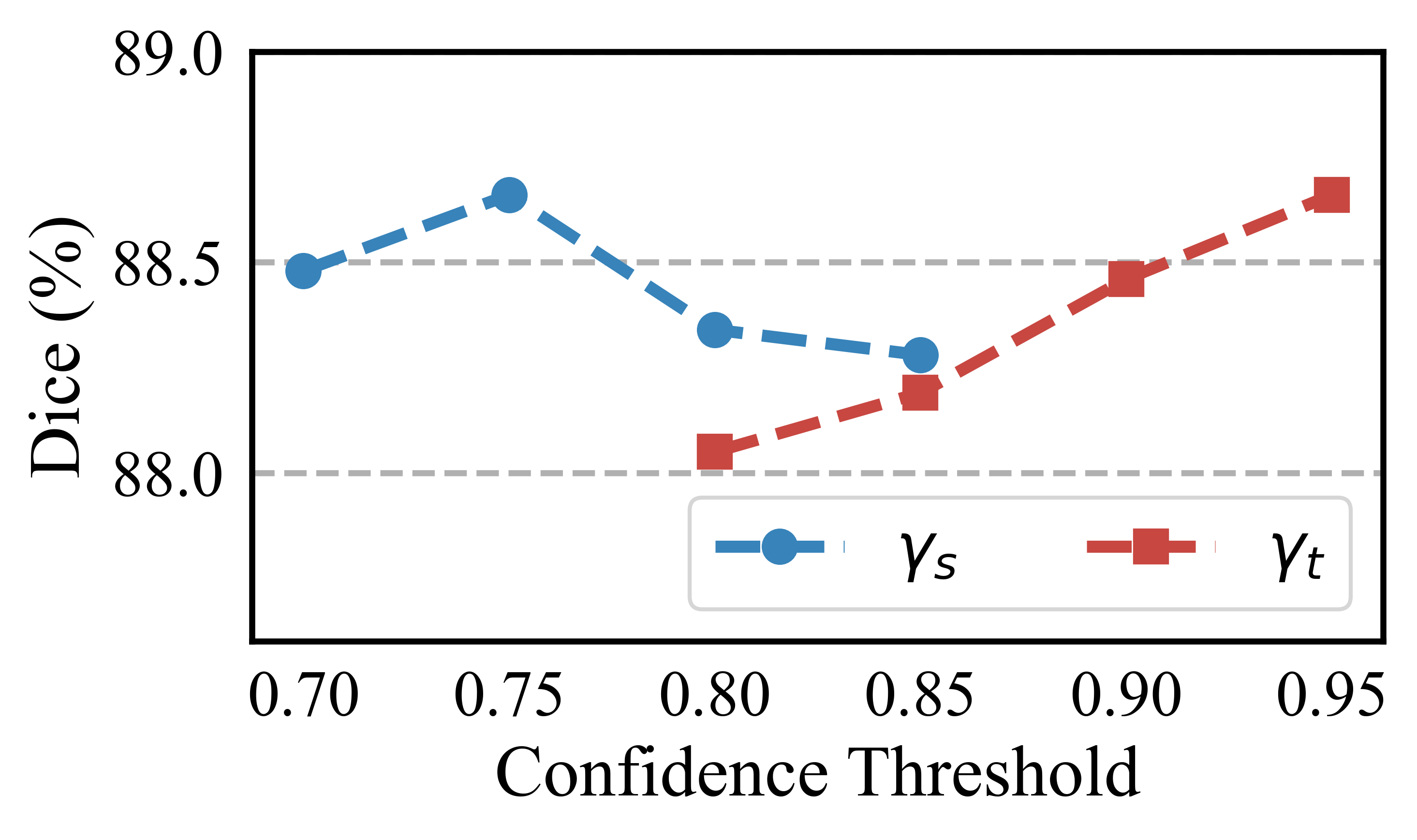}
  \caption{Influence of confidence thresholds $\gamma_s$ and $\gamma_t$ on ACDC dataset.} \label{hyper}
  \medskip
\end{minipage}
\label{fig:res}
\end{figure}

\begin{figure}[htbp]
\centering
\includegraphics[width=1\linewidth]{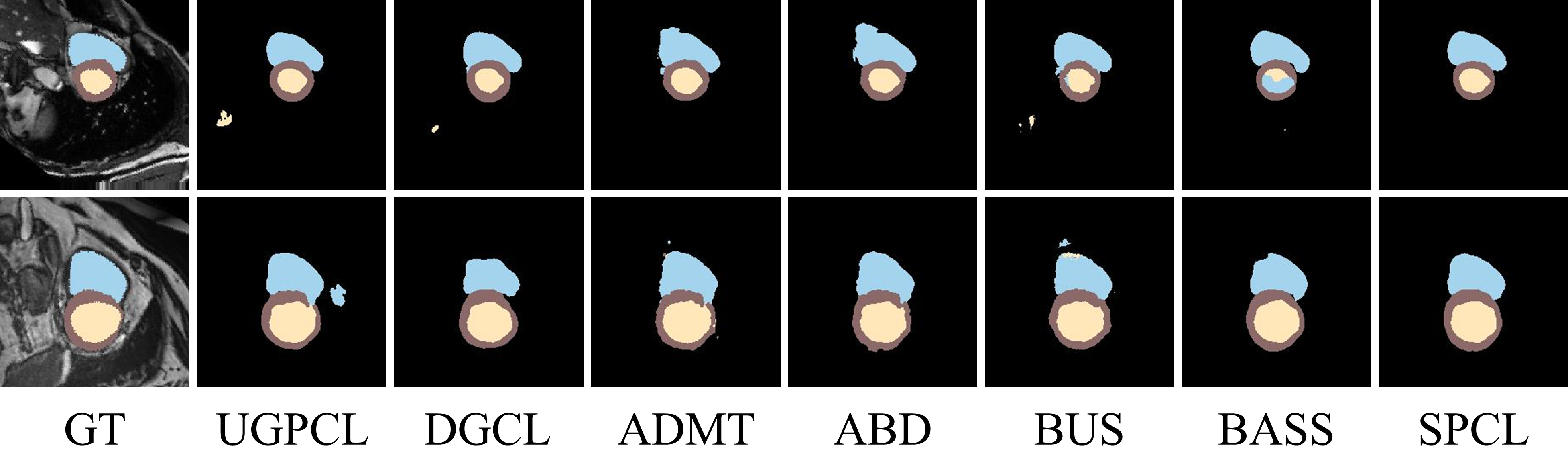}
\caption{Illustration of segmentation results using different methods.}
\label{sg}
\end{figure}

We compare our method with recent semi-supervised methods, including non-contrastive learning methods ADMT~\cite{zhao2024alternate} and ABD~\cite{chi2024adaptive}, contrastive learning methods  UGPCL~\cite{wang2022uncertainty} and DGCL~\cite{wang2023hunting}, as well as boundary-aware contrastive learning methods BUS~\cite{choe2024open} and BASS~\cite{zhang2024boundaryaware}.
Experimental results are listed in Table~\ref{whole-result}; our method outperforms these methods in all datasets and label ratios, particularly with smaller labeled ratios. 
It validates the effectiveness in subclass division and contrastive learning with unconcerned samples.
Fig.~\ref{tsne} illustrates the distributions of feature spaces produced by the three BACL methods, visualized using t-SNE. Our SPCL method demonstrates improved intra-class cohesion and enhanced inter-class separability.
Fig.~\ref{hyper} shows the influence of the confidence thresholds $\gamma_s$ and $\gamma_t$ on the performance, and $\gamma_s=0.75$ as well as $\gamma_t=0.95$ are set in experiments.
Moreover, Fig.~\ref{sg} demonstrates that our method preserves finer segmentation details compared to other methods.

\vspace{-0.3cm}

% p test
\begin{table}
  \caption{Statistical analysis of the SPCL and other semi-supervised learning methods on SCD, ACDC and M\&Ms datasets.}
  \scriptsize
  \label{p-test}
  \centering
  \renewcommand{\arraystretch}{1.2} 
  \setlength{\tabcolsep}{1.5mm}{
\begin{tabular}{lcccc}
\bottomrule
      & $Ranking$ & $Z_{value}$ & $P_{unadjusted}$ & $P_{Finner}$ \\
\hline
UGPCL~\cite{wang2022uncertainty} & 6.8333 & 4.677072 & 0.000003 & \textbf{0.000017} \\
DGCL~\cite{wang2023hunting}  & 4.8333 & 3.073504 & 0.002116 & \textbf{0.004227} \\
ADMT~\cite{zhao2024alternate}  & 5.6667 & 3.741657 & 0.000183 & \textbf{0.000548} \\
ABD~\cite{chi2024adaptive}   & 3.3333 & 1.870829 & 0.061369 & \textbf{0.073183} \\
BUS~\cite{choe2024open}    & 4.0000 & 2.405351 & 0.016157 & \textbf{0.024137} \\
BASS~\cite{zhang2024boundaryaware}  & 2.3333 & 1.069045 & 0.285049 & 0.285049 \\
SPCL (\textbf{Ours})  & \textbf{1.0000} & - & - & - \\
\toprule
\end{tabular}%
}
\end{table}

To evaluate the significance of differences between SPCL and other semi-supervised methods, we conducted the Friedman test~\cite{sheldon1996use} at a significance level of 0.1, using $Z_{value}$ to obtain probabilities from the normal distribution. Finner's procedure was applied to adjust $P$-values, accounting for multiple comparisons. The results are summarized in Table~\ref{p-test}, where $P_{unadjusted}$ and $P_{Finner}$ denote values before and after adjustment. SPCL achieves the highest ranking and, based on $P_{Finner}$, is significantly superior to all other methods except BASS. 

% component ablation
\begin{table}[htbp]
  \caption{Ablation study for the key components of our SPCL on the ACDC dataset.}
  \scriptsize
  \label{ab-result}
  \centering
   \renewcommand{\arraystretch}{1.1} 
 \setlength{\tabcolsep}{2.0mm}
\begin{tabular}{ccccc|ccc}
\bottomrule
\multicolumn{5}{c|}{Components}       & \multicolumn{3}{c}{ACDC$_{5\%}$} \\
\hline
$\mathcal{L}_{sup}$  & $\mathcal{L}_{unsup}$ & $\mathcal{L}_{icl}$  & $\mathcal{L}_{bcl}$  & $\mathcal{L}_{spcl}$ & Dice$\uparrow$  & Jacc$\uparrow$ & HD95$\downarrow$ \\
\hline
 \checkmark &       &       &       &       & 77.52 & 65.38 & 26.87 \\
 \checkmark &  \checkmark &       &       &       & 80.41 & 69.44 & 3.81 \\
 \checkmark &  \checkmark &  \checkmark &       &       & 87.77 & 78.92 & 3.20 \\
 \checkmark &  \checkmark &       &  \checkmark &       & 85.54 & 76.19 & 3.58 \\
 \checkmark &  \checkmark &       &       &  \checkmark & \textbf{88.66} & \textbf{80.21} & \textbf{2.81} \\
\toprule
\end{tabular}%
\end{table}

\begin{figure}[htb]
\begin{minipage}[b]{.38\linewidth}
  \centering
    \captionof{table}{Ablation study for subclass relation.}  \label{ab-rel}
  \begin{tabular}{c|c}
    \toprule
    Relation & Dice$\uparrow$ \\ \hline
    \textit{Neg}   & 87.20\\
    \textit{Pos}   & 88.18\\
    \textit{US} & \textbf{88.66}\\ 
    \bottomrule
  \end{tabular}
  \medskip
\end{minipage}
\hfill
\begin{minipage}[b]{0.6\linewidth}
  \centering
  \includegraphics[width=5.2cm]{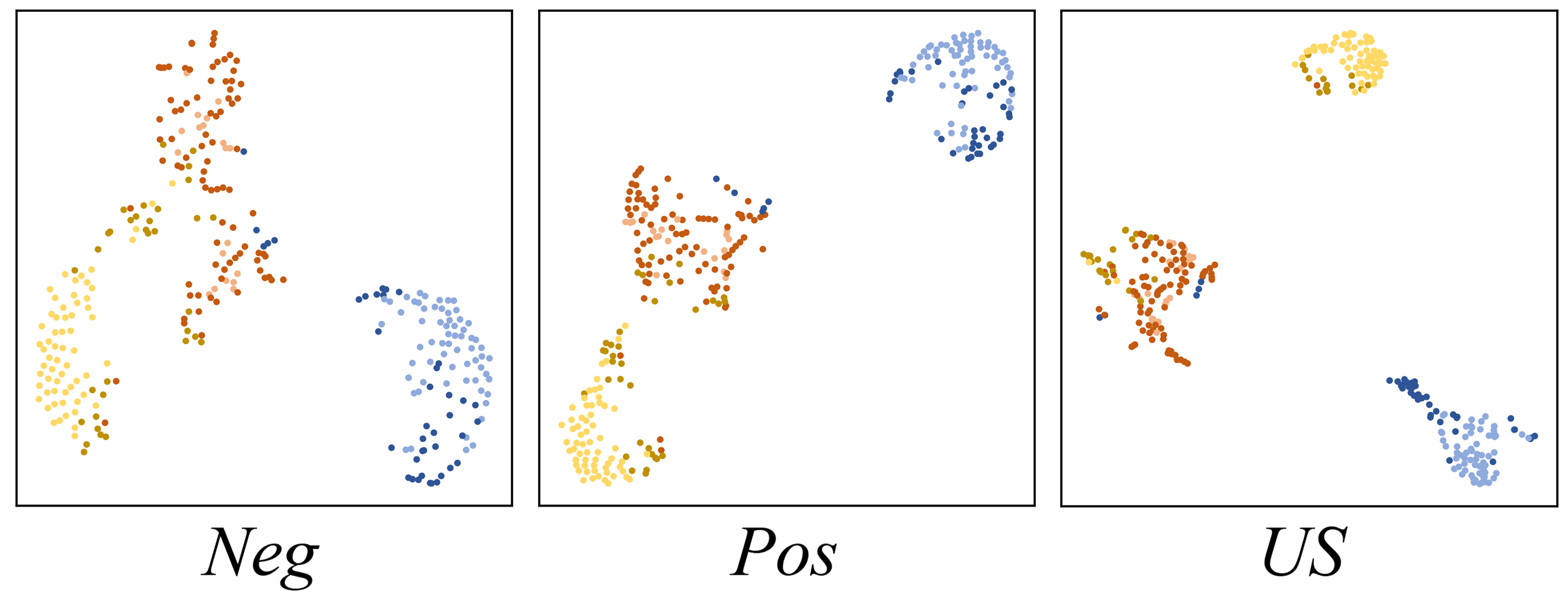}
  \caption{Visualization of feature space using different subclass relations.} \label{tsne-rel}
  \medskip
\end{minipage}
\label{fig:res2}
\end{figure}

\subsection{Ablation Study}
We first performed ablation studies to analyze step-by-step some critical components of our method on the ACDC dataset. As shown in Table~\ref{ab-result}, the unsupervised loss $\mathcal{L}_{unsup}$ enables the segmentation model to mine effective supervision from unlabeled data, improving the performance. Additionally, incorporating our inner contrast loss $\mathcal{L}_{icl}$ or our boundary contrast loss  $\mathcal{L}_{bcl}$ can effectively regularize the representation space, resulting in a significant performance improvement. 
Then, we explore the relationship between subclasses on the ACDC dataset, as shown in Table~\ref{ab-rel} and Fig.~\ref{tsne-rel}. \textit{Pos} indicates that the inner and boundary subclasses serve as positive samples for each other, while \textit{Neg} means that they act as negatives. \textit{US} denotes the use of unconcerned samples, which yields the best segmentation performance and regularizes a more compact and better-separated feature space.
Finally, our BCL loss shows a 0.45\% improvement in the Dice score (88.66 vs. 88.21) compared to the InfoNCE loss, demonstrating the effectiveness of our BCL on the boundary area.

\section{Conclusion} \label{conclusion}
This paper establishes a novel perspective on contrastive learning for cardiac image segmentation by explicitly introducing the intra-class subdivision and boundary-aware mechanism. By compacting intra-class features through “unconcerned sample” and improving boundary discrimination through a dedicated contrastive loss, the proposed SPCL framework addresses the challenges of feature heterogeneity and boundary ambiguity in medical imaging. Beyond achieving state-of-the-art performance under limited supervision, our approach highlights the importance of modeling fine-grained subclass structures in contrastive learning. Future work will explore more flexible subclass definitions, leveraging feature-space clustering to guide subclass partitioning for improved contrastive learning.

\clearpage
% -------------------------------------------------------------------------
\bibliographystyle{IEEEbib}
\bibliography{reference}  

\end{document}